\documentclass[10pt,conference,a4paper]{IEEEtran}
\usepackage{times,amsmath,epsfig}
\usepackage{epstopdf}

\usepackage{algorithm}
\usepackage{algpseudocode}
\usepackage{subfloat}
\usepackage{subfig}
\usepackage{overpic}
\usepackage{amssymb}
\usepackage{mathtools}
\usepackage{color}
\usepackage{multirow}

\floatname{algorithm}{Kernel}
\floatname{deviceAlgorithm}{Device function}

\title{An Optimized Union-Find Algorithm for Connected Components Labeling Using GPUs}
\author{%
{Jun Chen{\small $~^{*}$}, Qiang Yao, Houari Sabirin, Keisuke Nonaka, Hiroshi Sankoh, Sei Naito}
\vspace{1.6mm}\\
\fontsize{10}{10}\selectfont\itshape
\,Ultra-realistic Communication Group, KDDI Research, Inc.\\ 
Ohara 2-1-15, Fujimimo, Saitama, Japan
\fontsize{9}{9}\selectfont\ttfamily\upshape
\vspace{1.2mm}\\
\fontsize{10}{10}\selectfont\rmfamily\itshape
\,\{ju-chen, qi-yao, ho-sabirin, ki-nonaka, sankoh, sei\}@kddi-research.jp\\ 

\fontsize{9}{9}\selectfont\ttfamily\upshape
\,
}
\begin{document}
\maketitle

% INCLUDES COPYRIGHT NOTICE: one of three copyright notice should be included. Uncomment the appropriate line below, according to the authors affiliation:
\begin{figure}[b]
\parbox{\hsize}{\em
%information about the event:
%IEEE VCIP'14, Dec. 7 - Dec. 10, 2014, Valletta, Malta.

%copyright notice: one of three copyright notices below should be included. Uncomment the appropriate line, according to the authors affiliation:
%000-0-0000-0000-0/00/\$31.00 \ \copyright 2014 IEEE.
%U.S. Government work not protected by U.S. copyright.
%???-?-????-????-?/10/\$??.?? \copyright 2014 Crown.
}\end{figure}

\begin{abstract}
In this paper, we report on an optimized union-find (UF) algorithm that can label the connected components on a 2D image efficiently by employing GPU architecture. 
The proposed method comprises three phases: UF-based local merge, boundary analysis, and link. 
The coarse labeling in local merge, which makes computation efficient because the length of the label-equivalence list is sharply suppressed .
Boundary analysis only manages the cells on the boundary of each thread block to launch fewer CUDA threads. 
We compared our method with the label equivalence algorithm \cite{hawick2010parallel}, conventional parallel UF algorithm \cite{oliveira2010study}, and line-based UF algorithm \cite{yonehara2015line}.
Evaluation results show that the proposed algorithm speeds up the average running time by around $5$x, $3$x, and $1.3$x, respectively.
\\[1\baselineskip]
\end{abstract}

% NOTE keywords are not used for conference papers so do not populate them
%\begin{keywords}
%Image processing, Signal processing, Connected components labeling, Parallel computing, GPU and CUDA
%\end{keywords}
%

\section{Introduction}
Connected components labeling (CCL) is a task to give a unique ID to each connected region in a 2D/3D grid, which means that the input data is divided into separate groups where the elements from a single group share the same ID.
As a basic data clustering method, it is employed in numerous research areas like image processing, computer vision, and visual communication \cite{CCLinCUDA}.
W. Song, et al. \cite{song2016motion} presented a motion based skin region of interest detection method using a real-time CCL algorithm to reduce its execution time. 
A fast 3D shape measurement technique using blink-dot projection patterns that utilizes a CCL algorithm to compute the size and location of each dot on the captured images has been reported \cite{chen2013fast} \cite{chen2015blink}.  
P. Guler, et al. proposed a real-time multi-camera video analytics system \cite{guler2016real} employing CCL to perform noise reduction.
%CPU method & GPU method

%The real-time property of a CCL algorithm is very important for many systems and algorithms in acting as a fundamental operation.
%Numerous researches regarding fast sequential CCL, such as cell-based CCL \cite{qingyi2012fast} and CCL using binary decision tree \cite{chang2015block}, have been presented. 
%However, most of these approaches are suitable for sequential processing but not work efficiently on the parallel devices. 
{On the basis of the fact that parallel devices find countless applications in both industrial and academic areas, some CCL algorithms using GPUs have emerged \cite{iverson2015evaluation} recently to improve the real-time property of CCL, which is very important for many applications.
%Tab.~\ref{tab:Classification} shows the two classifications of parallel CCL algorithms, multi-pass method and one-pass method, according to whether they apply a convergence criterion or not.
The CCL algorithms can be classified into two categories, the multi-pass method and one-pass method, according to whether they apply a convergence criterion or not \cite{He201725}.
Tab.~\ref{tab:Classification} summarizes five typical parallel CCL approaches and a brief explanation is given in the following.} 
Neighbor propagation \cite{hawick2010parallel} is the simplest multi-pass approach that scans the neighborhood of a target cell to get the lowest label of a neighboring cell belonging to the same group. 
Row-column unification \cite{kalentev2011connected} enlarges the scan scope by allocating one row to each thread.
Label equivalence \cite{hawick2010parallel} employs neighbor propagation as the first phase to construct label-equivalence chains, and the following analysis and relabeling phases find the roots of each chain. 
The resolution of an input image determines the iteration times of neighbor propagation, while the iteration of row-column unification and label equivalence depend on the complexity of an input image.
The usual union-find (UF) algorithm \cite{introduction2Algo} is parallelized by dividing the input image into independent 2D blocks; local merge and global merge are introduced to solve the connectivity \cite{oliveira2010study}. 
Instead of using 2D blocks, a line-based parallel UF algorithm \cite{yonehara2015line} collects the pixels in one row to perform local label unification.
Even the computation of each kernel in such one-pass methods is heavier than those of multi-pass approaches; they label an image faster because each kernel only runs one time.

%
%\begin{table}[t]
%\centering
%    \caption{Classification of the parallel CCL algorithms}     % NOTE!  caption goes _before_ the table contents !!
%    \label{tab:Classification}
%	\centering
%    \begin{small}
%	\begin{tabular}{ |c|c| }
%	\hline
%	\multicolumn{3}{ |c| }{Team sheet} \\
%	\hline
%	Multi-pass methods &  One-pass methods \\ 
%	\hline
%	\multirow{3}{*}
%	{\!\!\!\!\! lena \!\!\! $(\! 512 \! \times \! 512 \!)$ \!\!\!\!\!} 
% 	 {Neighbour Propagation \cite{hawick2010parallel}} & {Conventional UF \cite{oliveira2010study}} \\
%	 {Row-Column Unification \cite{kalentev2011connected}} & {Line-based UF \cite{yonehara2015line}}  \\
% 	 {Label Equivalence \cite{hawick2010parallel} \cite{kalentev2011connected}} &  \\ 
% 	\hline  
%	\end{tabular}
%    %\end{small}
%\end{table}

\begin{table}[t]
\centering
    \caption{Classification of the parallel CCL algorithms}     % NOTE!  caption goes _before_ the table contents !!
    \label{tab:Classification}
	\centering
    %\begin{small}
	\begin{tabular}{ |c|c|c| }
	%\hline
	%\multicolumn{3}{ |c| }{Team sheet} \\
	\hline
	Method &  Scan Mode & Computational cost\\ 
	\hline
	%\multirow{3}{*}
	%{\!\!\!\!\! lena \!\!\! $(\! 512 \! \times \! 512 \!)$ \!\!\!\!\!} 
	{Neighbour Propagation \cite{hawick2010parallel}} & {Multi-Pass} & {High}\\
	{Row-Column Unification \cite{kalentev2011connected}} & {Multi-Pass} & {High}\\
	{Label Equivalence \cite{hawick2010parallel}} & {Multi-Pass} & {High}\\
	{Conventional UF \cite{oliveira2010study}} & {One-Pass} & {Low}\\
	{Line-based UF \cite{yonehara2015line}} & {One-Pass} & {Low}\\
 	\hline  
	\end{tabular}
    %\end{small}
\end{table}

{{In this study, we propose an optimized UF algorithm that is an improved version of conventional parallel UF \cite{oliveira2010study} with an optimized local merge and lightweight boundary analysis.
Its concepts are: (1) row-column unification is performed using shared memory before local UF to reduce the complexity of an initialized local label map; (2) connectivity analysis is conducted only for the cells on the block boundary to decrease the number of required CUDA threads.
Compared with the conventional UF \cite{oliveira2010study}, our proposed approach can perform local merge more efficiently because the label-equivalence chains are extensively suppressed as a result of the coarse labeling.
%, which means the iteration times for local labelling is reduced sharply.
For the line-based UF \cite{yonehara2015line}, it can extract the local label map slightly faster than our method. However, its global merge phase takes much longer because global UF should be applied to all the cells in the input data.
}}

\section{Algorithm Description}
\label{sec:Algorithm}

%Our approach to solve fast CCL is an improved version of conventional parallel UF \cite{oliveira2010study} with an optimized local merge and a lightweight boundary analysis.
%Its concepts are: (1) a row-column unification is performed using shared memory before local UF to reduce the complexity of initialized local label map; (2) a connectivity analysis is conducted only for the cells on the block boundary to decrease the number of required CUDA threads.
%Compared with the conventional UF \cite{oliveira2010study}, our proposed approach can perform local merge more efficiently because the label-equivalence chains are suppressed extensively benefited by the coarse labeling, which means the iteration times for local labelling is reduced sharply.
%For the Line-based UF \cite{yonehara2015line}, it can extract the local label map a little bit faster than our method. However, its phase of global merge takes much longer time because global UF should be applied to all the cells in the input data.

In this section, we outline the three kernels of our method. 
In the first kernel, UF-based local merge, we perform a coarse labeling before finding the real root of each cell to reduce the computational complexity in each thread.
In the last two kernels, boundary analysis and link, we merge individual blocks together to generate a global label map.
%In the following sections, we describe the main steps of the algorithm.

\begin{algorithm}
	\caption{\textbf{Local UF merge with coarse labeling }}
	\label{Alg:LocalMerge}
	\begin{algorithmic}[1]
		\Require \text{Image} $I$ \text{of size} $N \times M$
		\Require \text{Both block dimension and grid dimension are 2D}		
		\Require $label_{sm}[],dBuff_{sm}[]$\text{ are on shared memory} 			        
		\Require $LabelMap[]$ \text{is on global memory}

		\State \textbf{declare} \text{int} $x,y,tid,temp,l,g_x,g_y,g_l$
		\State \textbf{declare} \text{int} $label_{sm}[],dBuff_{sm}[]$ 

		\State $x,y \gets$ \text{2D global thread id}			
		\State $\textbf{if} \ \  x < imgWidth \And y < imgHeight$
		\State \ \ \ \ $tid \gets$ \text{1D thread id within block}		
		\State \ \ \ \ $label_{sm}[tid] \gets tid$  
		\State \ \ \ \ $dBuff_{sm}[tid] \gets image[x,y]$  
		\State \ \ \ \ \text{call syncthreads()}
		
		\State \ \ \ \ \text{// row scan}
		\State \ \ \ \ $\textbf{if} \ \ dBuff_{sm}[tid] == dBuff_{sm}[tid-1]$ 
		\State \ \ \ \ \ \ \ \ $label_{sm}[tid] = label_{sm}[tid-1]$
		\State \ \ \ \ $\textbf{end if}$
		\State \ \ \ \ \text{call syncthreads()}
		
		\State \ \ \ \ \text{// column scan}
		\State \ \ \ \ $\textbf{if} \ \ dBuff_{sm}[tid] == dBuff_{sm}[tid-blockdim.x]$ 
		\State \ \ \ \ \ \ \ \ $label_{sm}[tid] \gets label_{sm}[tid--blockdim.x]$
		\State \ \ \ \ $\textbf{end if}$
		\State \ \ \ \ \text{call syncthreads()}
		
		\State \ \ \ \ \text{// row-column unification}
		\State \ \ \ \ $temp \gets tid$
		\State \ \ \ \ \textbf{while} $temp != label_{sm}[temp]$ 
		\State \ \ \ \ \ \ \ \ $temp \gets label_{sm}[temp]$ 
		\State \ \ \ \ \ \ \ \ $label_{sm}[tid] \gets temp$ 
		\State \ \ \ \ \textbf{end while}
		
		\State \ \ \ \ \text{// local union find}
		\State \ \ \ \ $\textbf{if} \ \ dBuff_{sm}[tid] == dBuff_{sm}[tid-1]$ 
		\State \ \ \ \ \ \ \ \ \text{findAndUnion}$(label_{sm}[], tid, tid - 1)$
		\State \ \ \ \ $\textbf{end if}$
		\State \ \ \ \ \text{call syncthreads()}
		\State \ \ \ \ $\textbf{if} \ \ dBuff_{sm}[tid] == dBuff_{sm}[tid-blockdim.x]$ 
		\State \ \ \ \ \ \ \ \ \text{findAndUnion}$(label_{sm}[], tid, tid-blockDim.x)$
		\State \ \ \ \ $\textbf{end if}$
		\State \ \ \ \ \text{call syncthreads()}
		
%		\State \ \ \ \ \text{// convert local index to global index}		
%		\State \ \ \ \ $l \gets find(label_{sm}[], tid)$
%		\State \ \ \ \ $l_x \gets l \ / \  blockdim.x$
%		\State \ \ \ \ $l_y \gets l \ \% \ blockdim.x$
%		\State \ \ \ \ $g_l \gets (blockIdx.x * blockDim.x + l_x) + (blockIdx.y * blockDim.y + l_y) * imgWidth$
%		\State \ \ \ \ $LabelMap[x,y] \gets g_l$
%		\State \ \ \ \ $LabelMap[x,y] = idConvert(l)$	
		
		\State \ \ \ \ \text{// convert local index to global index}		
		\State \ \ \ \ $l \gets find(label_{sm}[], tid)$
		\State \ \ \ \ $l_x \gets l \ / \  blockdim.x$
		\State \ \ \ \ $l_y \gets l \ \% \ blockdim.x$
		\State \ \ \ \ $g_l \gets (blockIdx.x * blockDim.x + l_x) + (blockIdx.y * blockDim.y + l_y) * imgWidth$
		\State \ \ \ \ $LabelMap[x,y] \gets g_l$	
		
		\State $\textbf{end if}$	
	\end{algorithmic}
\end{algorithm}

\begin{figure}[t]
\centering
\footnotesize
	\begin{minipage}[b]{0.48\linewidth}
 		\centering
 		\subfloat[input data]
		{
 	 		\begin{overpic}[width=1\textwidth]
 	 			{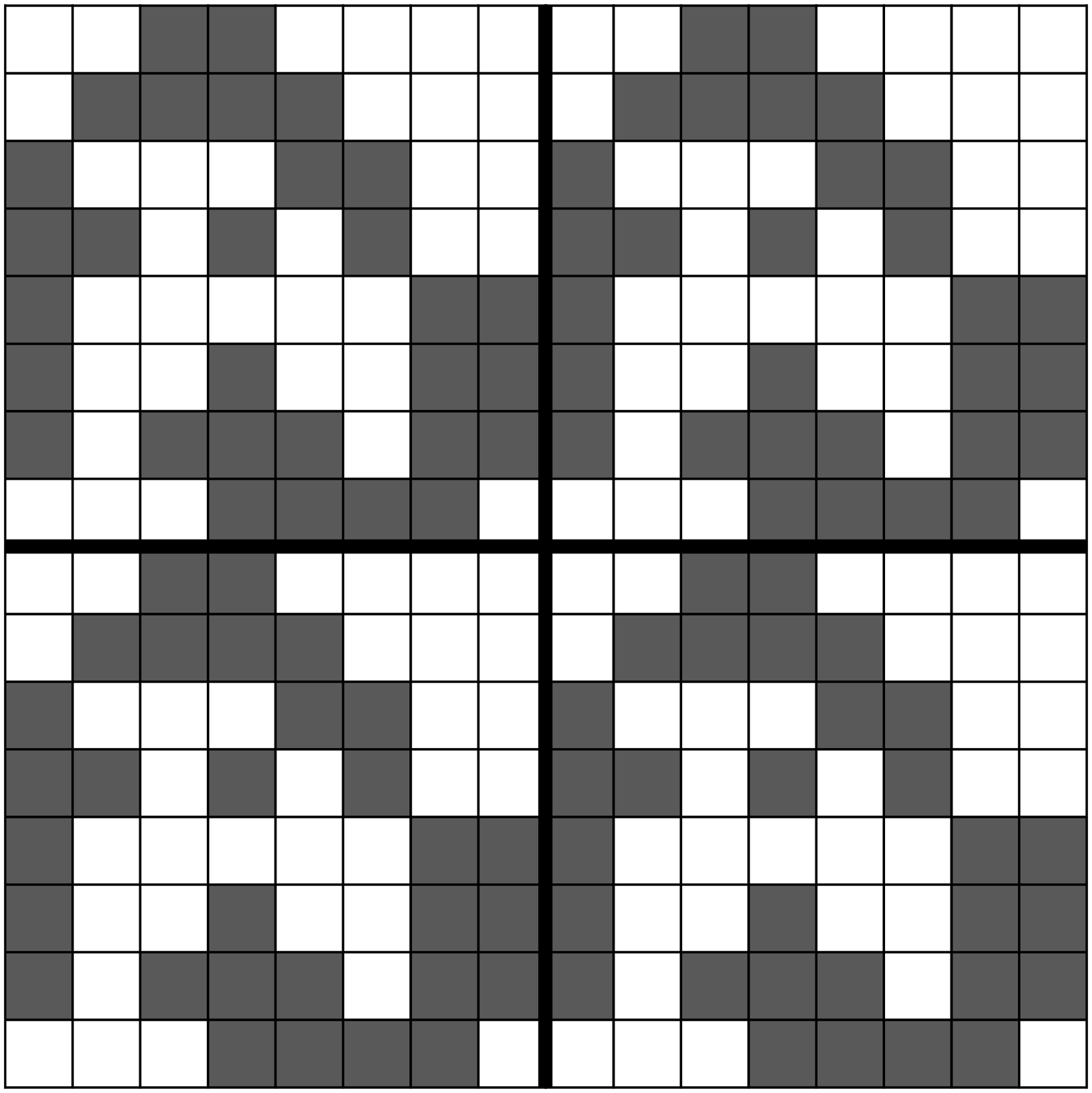}
 		\end{overpic}
 	 	}
	\end{minipage}
\hskip 2mm
	\begin{minipage}[b]{0.48\linewidth}
 		\centering
 		\subfloat[initialized local label map]
		{
 			\begin{overpic}[width=1\textwidth]
 	 			{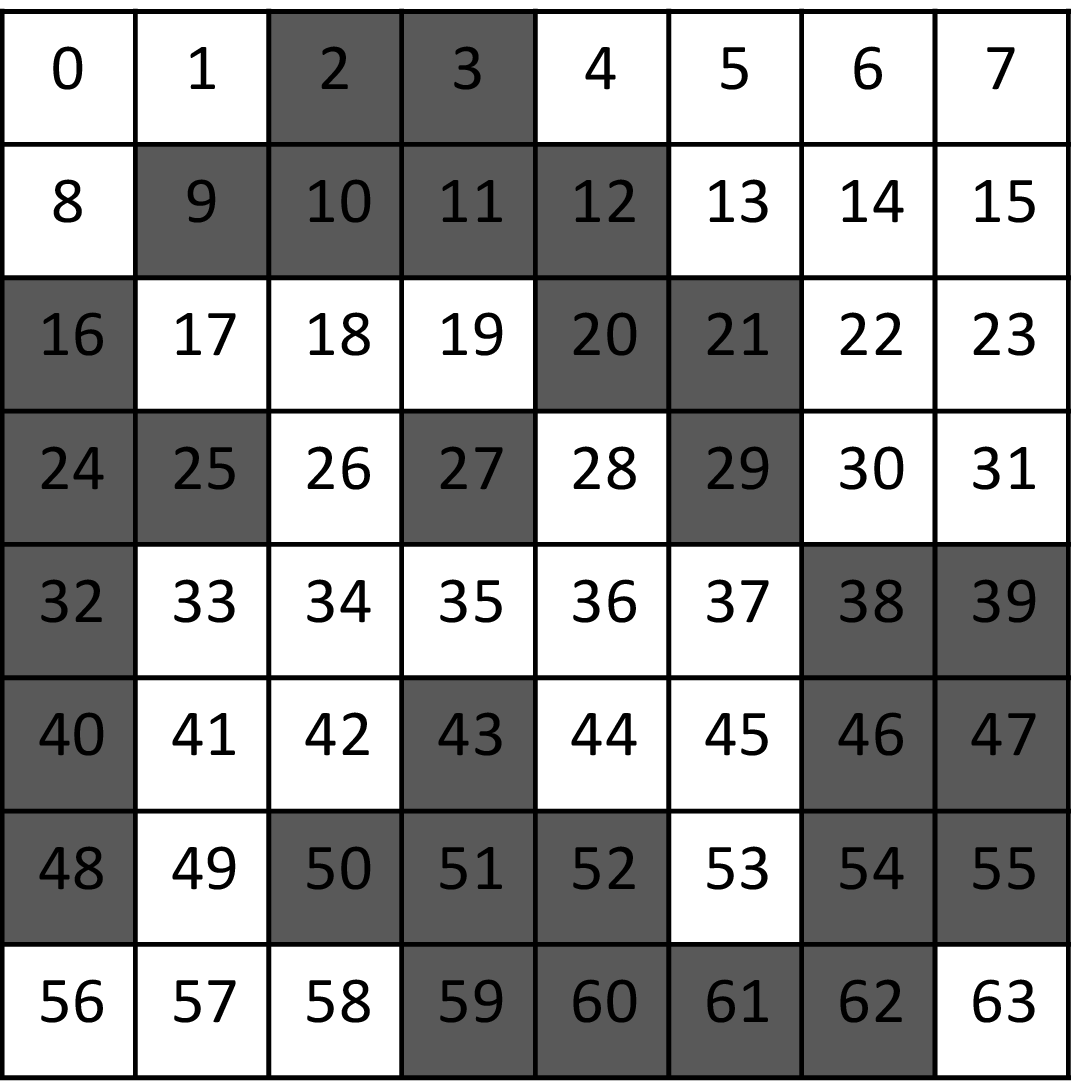}
 		\end{overpic}
 		}
	\end{minipage}
\caption{Input data and initialized local label map.}
\label{fig:initialization}
\vskip -3mm
\end{figure}

\begin{figure}[t]
\centering
\footnotesize
	\begin{minipage}[b]{0.47\linewidth}
 		\centering
 		\subfloat[row scan]
		{
 	 		\begin{overpic}[width=1\textwidth]
 	 			{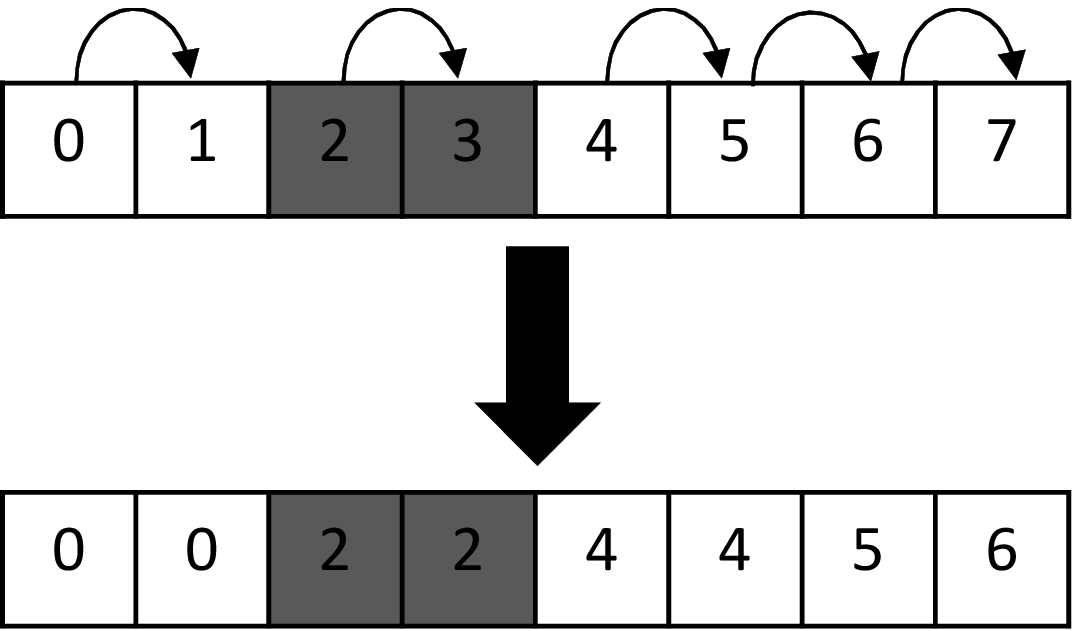}
 		\end{overpic}
 	 	}
	\end{minipage}
\hskip 2mm
	\begin{minipage}[b]{0.27\linewidth}
 		\centering
 		\subfloat[column scan]
		{
 			\begin{overpic}[width=1\textwidth]
 	 			{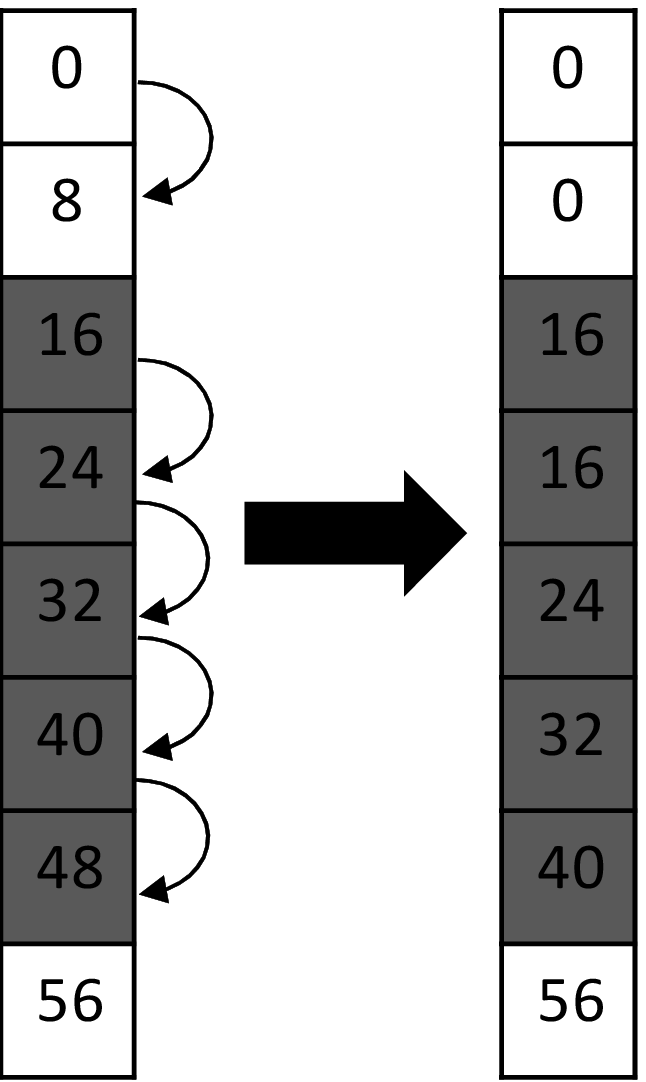}
 		\end{overpic}
 		}
	\end{minipage}
\caption{Scan model.}
\label{fig:ScanModel}
\vskip -3mm
\end{figure}

\subsection{Local merge with coarse labeling}
\label{sec:LocalMerge}

The first kernel, local merge with coarse labeling, consists of three phases: initialization, coarse labeling using row-column unification, and local UF.
Its pseudo-code is listed in Kernel~\ref{Alg:LocalMerge} by following 4-connectivity. 

\subsubsection{Initialization} 
\label{sec:initialization}
\vskip 1mm
\noindent

We divide the input image into several rectangular pieces, as shown in Fig.~\ref{fig:initialization} (a), and assign each piece to different GPU threads blocks where the threads can cooperate with each other using shared memory and can be synchronized \cite{nvidapg}.
The cells in each block are indexed by the thread ID within the block.  
Fig.~\ref{fig:initialization} (b) presents an example of an $8 \times 8$ initialized local label map that was allocated on shared memory. 
Here, the gray cells represent foreground areas, while the white cells represent background areas.

\subsubsection{Coarse labeling using row-column unification}
\label{sec:RowColumnUnification}
\vskip 1mm
\noindent

\begin{figure}[t]
\centering
\footnotesize
	\begin{minipage}[b]{0.47\linewidth}
 		\centering
 		\subfloat[local label map after row-column scan]
		{
 	 		\begin{overpic}[width=1\textwidth]
 	 			{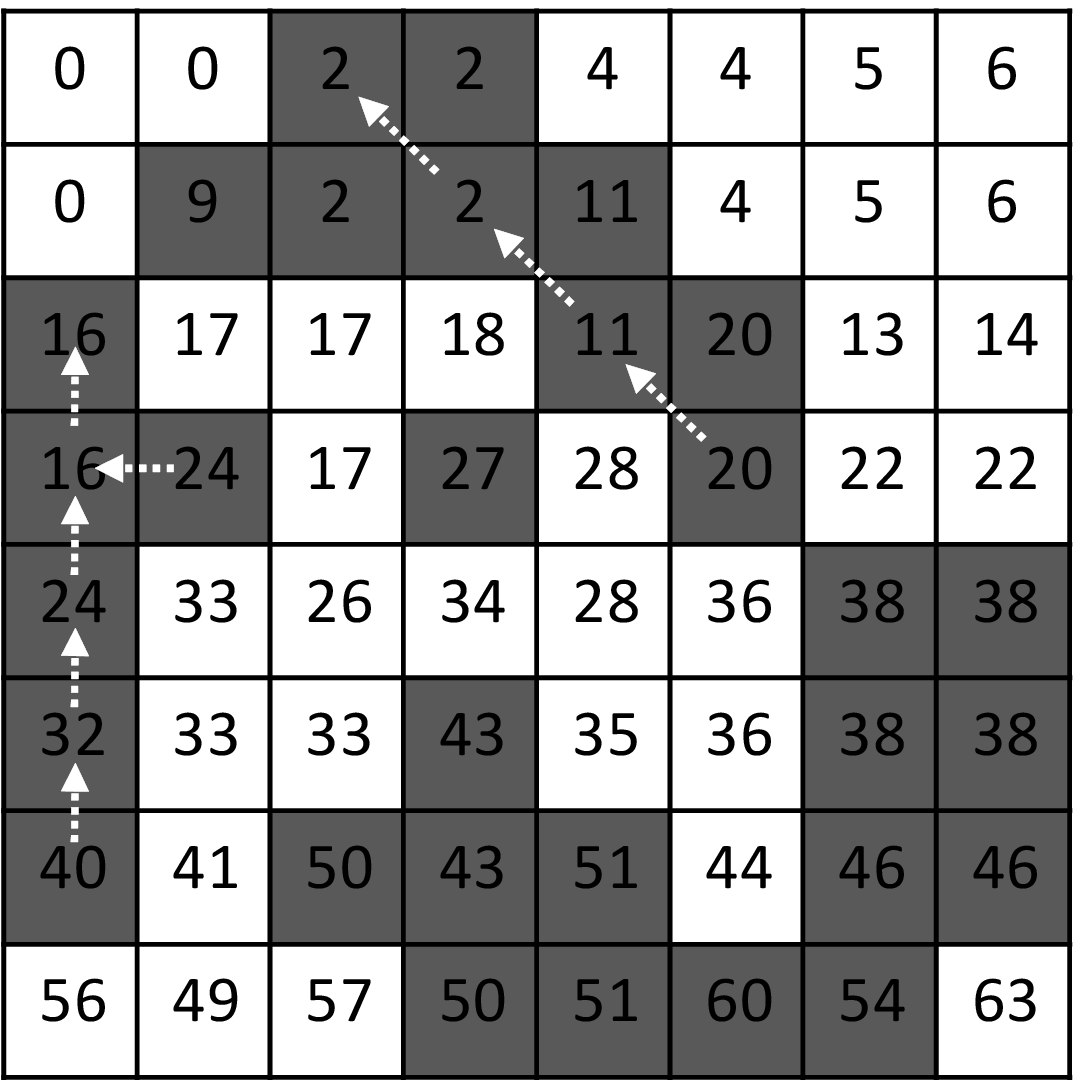}
 		\end{overpic}
 	 	}
	\end{minipage}
\hskip 2mm
	\begin{minipage}[b]{0.47\linewidth}
 		\centering
 		\subfloat[local label map after row-column unification]
		{
 			\begin{overpic}[width=1\textwidth]
 	 			{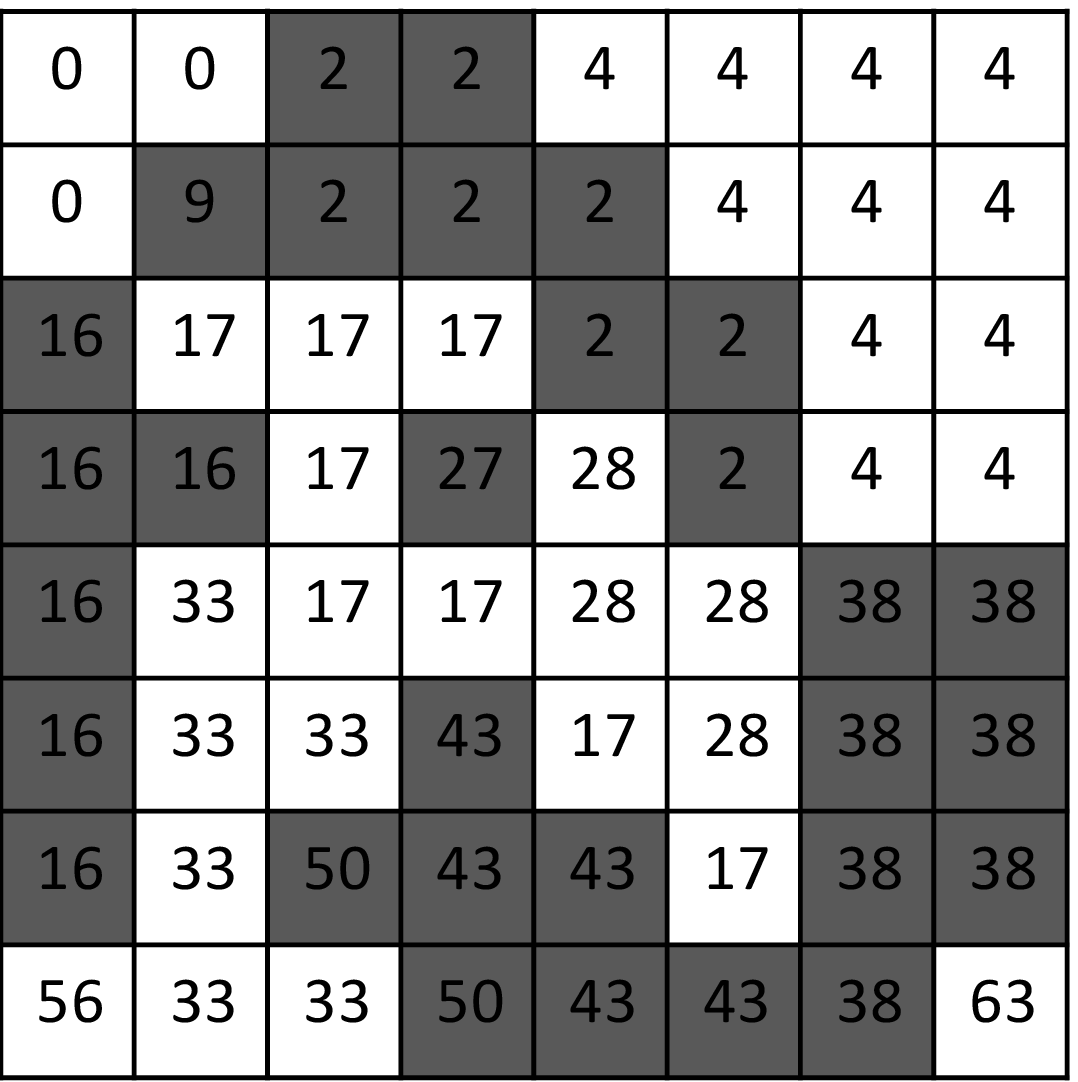}
 		\end{overpic}
 		}
	\end{minipage}
\caption{Coarse labeling.}
\label{fig:Coarselabeling}
\vskip -3mm
\end{figure}

In an initialized local label map, as shown in Fig.~\ref{fig:initialization} (b), the label of the left cell and the label of the upper cell are always smaller than that of a target cell, while the upper one is always smaller than the left one. 
Based on this fact, we scan the rows first and then go to column scan.
The cell will get the label of its neighboring cell, left or upper, with the same property. 
Fig.~\ref{fig:ScanModel} shows the scan models.
Unlike the methods that record the entire label-equivalence lists, this method records the lowest label that the label is equivalent to. 
Its memory access complexity is reduced due to the utilization of shared memory, while the equivalence can be unified by a low number of iteration because the dimension of a thread block is limited by the CUDA runtime system.
Fig.~\ref{fig:Coarselabeling} (a) presents two equivalence lists in the local label map after row-column scan.
Fig.~\ref{fig:Coarselabeling} (b) shows the coarsely labeled label map after row-column unification.

\begin{algorithm}
	\caption{\textbf{Boundary analysis}}
	\label{Alg:scanMatching}
	\begin{algorithmic}[1]
		\Require \text{Image} $I$ \text{of size} $N \times M$
		\Require \text{Both block dimension and grid dimension are 2D}		
		\Require $LabelMap[]$ \text{is on global memory}
		
		\State \textbf{declare} \text{int} $id,h_x,h_y,v_x,v_y, pInLine, p_h, p_v$
		\State \textbf{declare} \text{bool} $b_h, b_v$
		\State $id \gets$ \text{1D global thread id}
		
		\State \text{// convert 1D global thread id to 2D cell id}% on the }
		%\State \text{// block boundary along x-axis}
		\State $h_x \gets id \ \% \ imgWidth$
		\State $h_y \gets id \ / \ (imgWidth * blockDim.y)$
		
		%\State \text{// convert global thread ID to cell ID on the }
		%\State \text{// block boundary along y-axis}
		\State $pInLine \gets imgWidth \ / \ blockDim.x$
		\State $v_x \gets id \ \% \ pInLine * blockDim.x$
		\State $v_y \gets id \ / \ pInLine$

		\State $p_h \gets h_x + h_y * imgWidth$
		\State $p_v \gets v_x + v_y * imgWidth$
		
		\State $b_h \gets h_x < imgWidth \And h_y < imgHeight$
		\State $b_v \gets v_x < imgWidth \And v_y < imgHeight$
		
		\State \text{// boundary analysis along x-axis}
		\State $\textbf{if} \ \  b_h \And image[h_x,h_y] == image[h_x-1,h_y]$
		\State  \ \ \ \ $\text{findAndUnion}(LabelMap, \ p_h, \ p_h - 1)$;
		\State $\textbf{end if}$	
		
		\State \text{// boundary analysis along y-axis}
		\State $\textbf{if} \ \  b_v \And image[v_x,v_y] == image[v_x,v_y-imgWidth]$
		\State  \ \ \ \ $\text{findAndUnion}(LabelMap, \ p_v, \ p_v - imgWidth)$;		
		\State $\textbf{end if}$
	\end{algorithmic}
\end{algorithm}

\subsubsection{Local UF}
\label{sec:localUF}
\vskip 1mm
\noindent

UF, expressed by $findAndUnion$ in Kernel~\ref{Alg:LocalMerge}, is a data structure that divides a set of elements into a number of disjoint subsets by using $find$ and $merge$ operations.
The $find$ is an iterative search to extract the root of a label-equivalence list and return its label. 
The $merge$ is a unification to assign the root label to the elements belonging to the subset.
\cite{oliveira2010study} gives a detailed description of these two operations.
By comparing the initialized local label map (Fig.~\ref{fig:initialization} (b)) with the one after row-column unification (Fig.~\ref{fig:Coarselabeling} (b)), it can be noticed that the path to find the root of a label-equivalence chain is compressed sharply, which enables local UF to run efficiently.
The final step of this kernel is an ID conversion that converts the local index to a global index.  
Fig.~\ref{fig:Boundaryanalysis} (a) presents a converted global label map.
%\begin{algorithm}
%	\caption{\textbf{Union find}}
%	\label{Alg:scanMatching}
%	\begin{algorithmic}[1]		
%		\Require $Array[]$ \text{is on global memory or shared memory}
%		\Require $p$ \text{and} $q$ \text{are going to be merged together}
%		\Require $p$ \text{and} $q$ \text{share a parent in} $Array[]$
%		
%		\State \textbf{declare} \text{bool} $b$
%		\State $b \gets flase$
%		
%		\State \textbf{do} 
%		\State \ \ \ \ $p \gets find$ $\text{find}(Array[], \ p)$
%		\State \ \ \ \ $q \gets find$ $\text{find}(Array[], \ q)$
%		
%		\State  \ \ \ \ \textbf{if} $p < q$
%		\State  \ \ \ \ \ \ \ \ $\text{atomicMin(Array[q], p)}$;
%		\State  \ \ \ \ \textbf{else if}  $p > q$
%		\State  \ \ \ \ \ \ \ \ $\text{atomicMin(Array[p], q)}$;
%		\State  \ \ \ \ \textbf{else}
%		\State  \ \ \ \ \ \ \ \ $b \gets true$;
%		\State  \ \ \ \ \textbf{end if}
%
%		\State \textbf{While} $!b$ 		
%	\end{algorithmic}
%\end{algorithm}
%
%
%\begin{algorithm}
%	\caption{\textbf{Solve the label chain}}
%	\label{Alg:scanMatching}
%	
%	\begin{algorithmic}[1]	
%		\Require $p$ \text{is the target pixel}
%		\Require $Array[]$ \text{is on global memory or shared memory}	
%		\State \textbf{while} $p != Array[p]$ 
%		\State \ \ \ \ $p \gets Array[p]$ 
%		\State \textbf{end while}
%		
%		\State \textbf{return} $p$
%	\end{algorithmic}
%\end{algorithm}	

\subsection{Boundary analysis}
\label{sec:Analysis}

\begin{figure}[t]
\footnotesize
\centering
%\vskip 4cm
\includegraphics[width=0.95\linewidth]{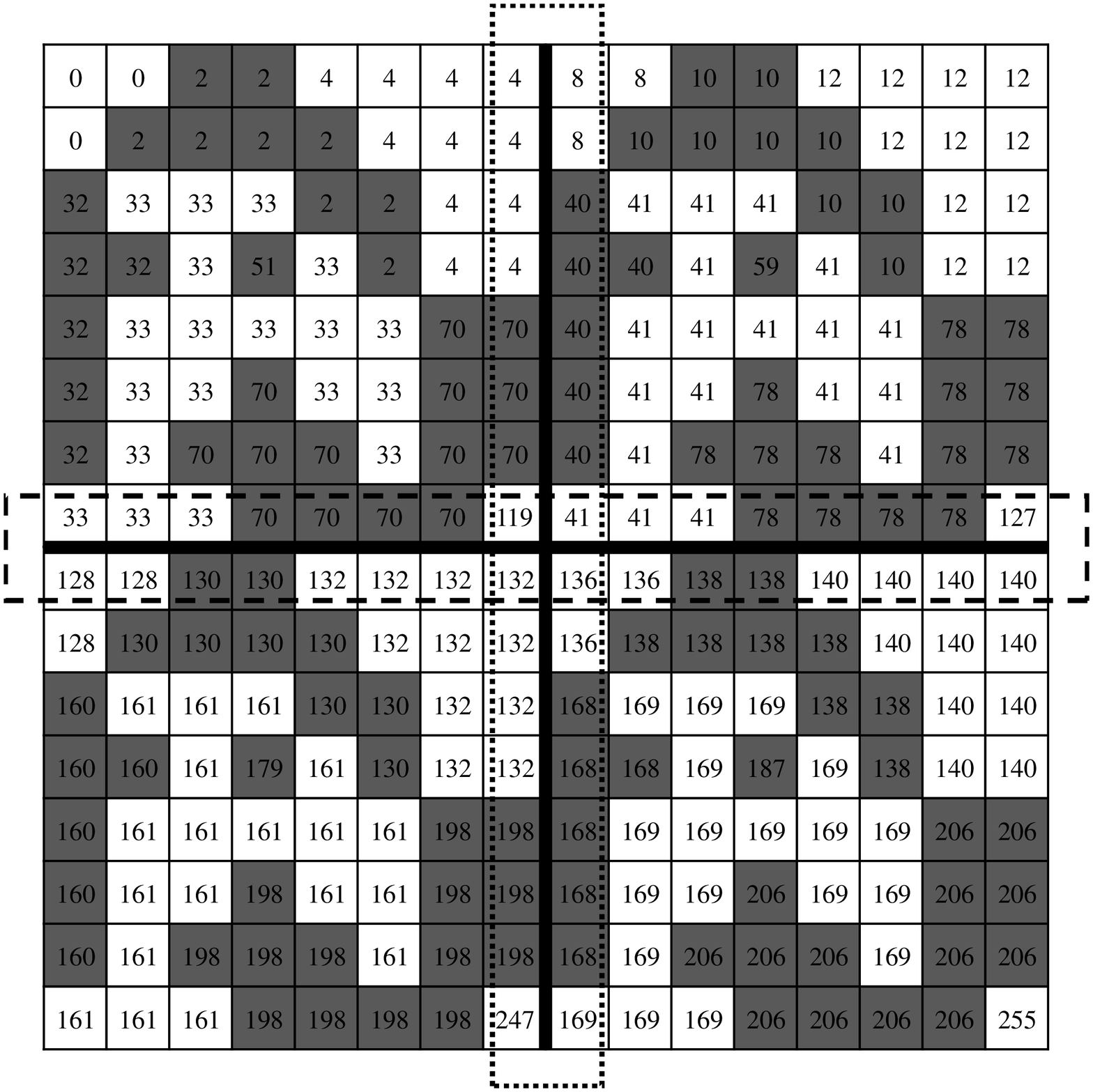}
\vskip 1mm
(a) global label map after local merge
\vskip 2mm

\includegraphics[width=0.95\linewidth]{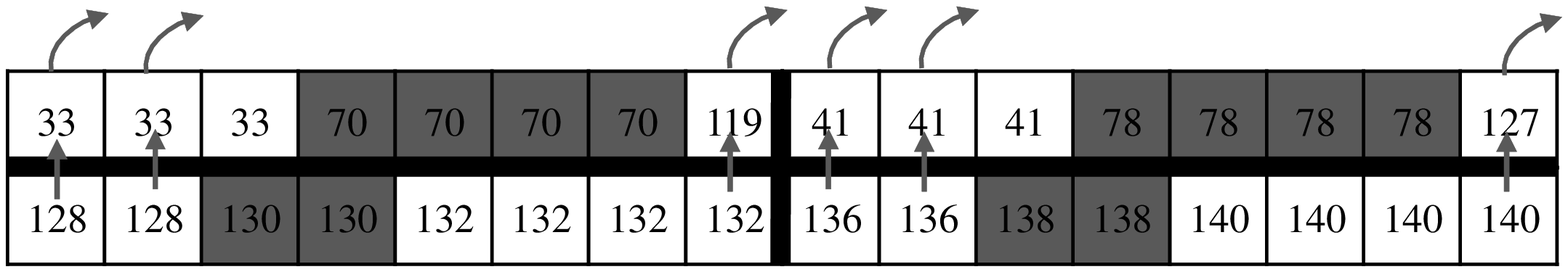}
\vskip 1mm
(b) boundary analysis along x-axis
%\vskip 2mm
\caption{Boundary analysis}
\label{fig:Boundaryanalysis}
%\vskip -1mm
\end{figure}

In the boundary analysis phase, we only perform UF for the cells on the block boundary (those marked on Fig.~\ref{fig:Boundaryanalysis} (a)) to launch fewer threads.
Assuming the resolution of an input image is $N \times M$ and the block configuration of the Kernel~\ref{Alg:LocalMerge} is $\{b_x,b_y,1\}$, the number of cells on the block boundary along $x$-axis and $y$-axis $\{P_x,P_y\}$ can be determined as follows:

\begin{eqnarray}
P_x = \left \lfloor N \ / \ b_x\right \rfloor  * M, \\
P_y = \left \lfloor M \ / \ b_y\right \rfloor  * N,
\end{eqnarray}

Here, $\left \lfloor x \right \rfloor$ means the largest integer smaller or equal to $x$. 
To integrate the boundary analysis along $x-$ and $y-$axis into one kernel, $\max\{P_x,P_y\}$ threads spawned by Kernel~\ref{Alg:scanMatching} should be invoked.  
Fig.~\ref{fig:Boundaryanalysis} (b) shows how to analyze the connectivity in the $x$-direction: the cell on the boundary merges with its upper cell by using UF if they have the same property.
The union along the $y$-direction works in the same manner.

\subsection{Final link}
\label{sec:link}

After analyzing the connectivities of the cells on the block boundary, the independent local label maps are associated as an entirety.
We compute the final global label map in the same way as that reported in \cite{oliveira2010study} and \cite{yonehara2015line}.

%\begin{algorithm}
%	\caption{\textbf{Link}}
%	\label{Alg:link}
%	\begin{algorithmic}[1]
%		\Require \text{Image} $I$ \text{of size} $N \times M$
%		\Require \text{Both block dimension and grid dimension are 2D}		
%		\Require $LabelMap[]$ \text{is on global memory}
%		
%		\State \textbf{declare} \text{int} $x,y, gid$
%		\State $x \gets blockIdx.x * blockDim.x + threadIdx.x$
%		\State $y \gets blockIdx.y * blockDim.y + threadIdx.y$
%		\State $gid \gets  x + y * imgWidth$		
%		
%		\State $\textbf{if} \ \  x < imgWidth \And y < imgHeight$
%		\State \ \ \ \ $LabelMap[x,y] \gets$ $\text{find}(LabelMap[], \ gid)$
%		\State $\textbf{end if}$	
%		
%	\end{algorithmic}
%\end{algorithm}

\section{Evaluation Experiments}
\label{sec:Evaluation}

\begin{figure}[t]
\centering
\footnotesize
	\begin{minipage}[b]{0.48\linewidth}
 		\centering
 		\subfloat[lena]
		{
 	 		\begin{overpic}[width=1\textwidth]
 	 			{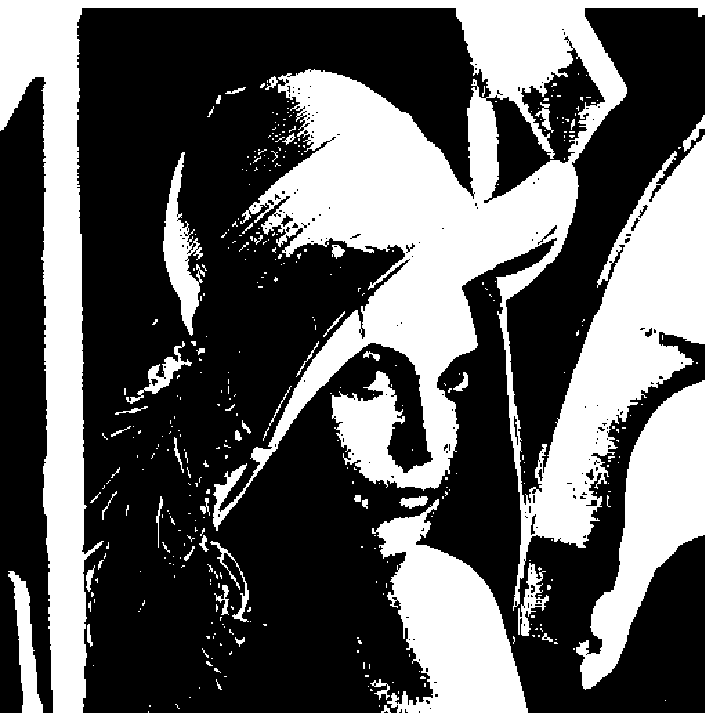}
 		\end{overpic}
 	 	}
	\end{minipage}
\hskip 2mm
	\begin{minipage}[b]{0.48\linewidth}
 		\centering
 		\subfloat[peppers]
		{
 			\begin{overpic}[width=1\textwidth]
 	 			{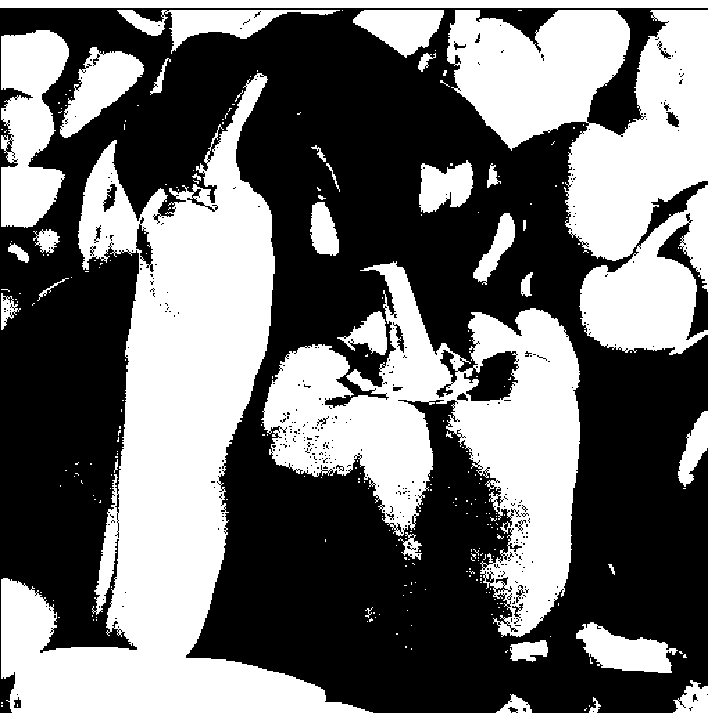}
 		\end{overpic}
 		}
	\end{minipage}
\caption{Test images.}
\label{fig:EvaluationSrc}
%\vskip -3mm
\end{figure}

%\begin{figure}[t]
%\footnotesize
%\centering
%%\vskip 4cm
%\includegraphics[width=0.95\linewidth]{fig/LenaTime.eps}
%\vskip 0.1mm
%(a) execution times of Lena
%\vskip 2mm
%
%\includegraphics[width=0.95\linewidth]{fig/PeppersTime.eps}
%\vskip 0.1mm
%(b) execution times of peppers
%%\vskip 2mm
%\caption{Time evaluation}
%\label{fig:EvaluationTime}
%%\vskip -1mm
%\end{figure}%
%
%
%
%\begin{table}[!h]
%\centering
%
%    \caption{Speedup ratio for different images}     % NOTE!  caption goes _before_ the table contents !!
%    \label{tab:speedup}
%	\centering
%
%    \begin{small}
%    \begin{tabular}{|l|c|c|c|c|}
%    \hline
%%    {\bfseries Images} & \multicolumn{3} {c|} {\bfseries Appearance (in Times New Roman or Times} \\
%%    \cline{2-4}
%    {\bfseries Images} & {\bfseries  LE} & {\bfseries UF}  & {\bfseries Line UF}  & {\bfseries proposed} \\
%    \hline
%    Lena $(512 \times 512) $        & 5.30	&	1.92	& 1.37	  & 1.00	\\
%    \hline
%    Lena $(1024 \times 1024) $      & 4.52	&	2.48	& 1.30	  & 1.00	\\
%    \hline
%    Lena $(2048 \times 2048) $      & 5.27	&	2.96	& 1.30	  & 1.00   \\
%    \hline
%    Lena $(4096 \times 4096) $      & 5.54	&	3.45	& 1.25	  & 1.00	\\
%    \hline
%    Peppers $(512 \times 512) $     & 6.55	&	1.93	& 1.33	  & 1.00	\\
%    \hline
%    Peppers $(1024 \times 1024) $   & 5.27	&	2.47	& 1.27	  & 1.00	\\
%    \hline
%    Peppers $(2048 \times 2048) $   & 5.76	&	3.08	& 1.29	  & 1.00   \\
%    \hline
%    Peppers $(4096 \times 4096) $   & 5.53	&	3.43	& 1.24	  & 1.00	\\
%    \hline
%    \end{tabular}
%    \end{small}
%\end{table}

To demonstrate the effectiveness of our proposed algorithm, we run it and the other three parallel methods, label equivalence (LE) \cite{CCLinCUDA}, conventional parallel UF \cite{oliveira2010study}, and line-based UF \cite{yonehara2015line}, on a PC equipped with an NVIDIA Geforce GTX 1070 for the images shown in Fig.~\ref{fig:EvaluationSrc}. 
For the line-based UF method, its thread blocks are configured as $\left \{ 512, 1, 1 \right \}$, while the configuration of the other three methods is $\left \{ 32, 16, 1 \right \}$.

Tab.~\ref{tab:speedup} shows the comparison results for the execution times of these algorithms with images of different size.
Here, we run each algorithm 100 times and take their extreme value as well as average value.
It can be seen that the optimized UF can label a $512 \times 512$, $1024 \times 1024$, $2048 \times 2048$, and $4096 \times 4096$ binary image in  around $0.14$, $0.40$, $1.10$, and $3.40$ ms respectively, while the other methods take longer to accomplish CCL.
%Tab.~\ref{tab:speedup} presents the speedup ratio for the same images.
From an analysis of these results, it can be deduced that the one-scan methods, UF, line-based UF, and our proposed method, work more efficiently than LE, one of the typical multi-scan methods.
Meanwhile, it indicates that our method outperforms the other two methods.
Compared with UF, the speedup ratio increases with the increase in image resolution. 
It is about $3.4$ times faster for a $4096 \times 4096$ binary image.
For the line-based UF, the speedup ratio is quite stable and is around $1.30$ for all the images.

\begin{table}[t]
\centering

    \caption{Execution time in millisecond for different images}     % NOTE!  caption goes _before_ the table contents !!
    \label{tab:speedup}
	\centering

    \begin{small}
	\begin{tabular}{ |c|c|c|c|c|c| }
	%\hline
	%\multicolumn{3}{ |c| }{Team sheet} \\
	\hline
	Images &  & {LE} & {UF} & {Line \!\!\! UF} & {ours} \\ 
	\hline
	\multirow{3}{*}
	{\!\!\!\!\! lena \!\!\! $(\! 512 \! \times \! 512 \!)$ \!\!\!\!\!} 
 	& min & 0.66 & 0.26 & 0.17 & 0.13\\
	& max & 1.03 & 0.39 & 0.44 & 0.18 \\
 	& mean& 0.73 & 0.28 & 0.19 & 0.14 \\ 
 	\hline
	\multirow{3}{*}
	{\!\!\!\!\! lena \!\!\! $(\!1024 \! \times \! 1024 \!)$ \!\!\!\!\!} 
 	& min & 1.61 & 0.96 & 0.49 & 0.38 \\
 	& max & 2.11 & 1.00 & 0.56 & 0.49 \\
 	& mean & 1.77 & 0.97 & 0.51 & 0.40 \\ 
 	\hline
	\multirow{3}{*}
	{\!\!\!\!\! lena \!\!\! $(\!2048 \! \times \! 2048 \!)$ \!\!\!\!\!} 
 	& min & 5.13 & 2.97 & 1.29 & 0.99 \\
 	& max & 5.65 & 3.05 & 1.45 & 1.11 \\
 	& mean & 5.38 & 2.99 & 1.32 & 1.02 \\ 
 	\hline
	\multirow{3}{*}
	{\!\!\!\!\! lena \!\!\! $(\!4096 \! \times \! 4096 \!)$ \!\!\!\!\!} 
 	& min & 18.40 & 11.49 & 4.16 & 3.30 \\
 	& max & 19.08 & 11.71 & 4.40 & 4.15 \\
 	& mean & 18.64 & 11.56 & 4.21 & 3.36 \\ 
 	\hline
	\multirow{3}{*}
	{\!\!\!\!\! peppers \!\!\! $(\! 512 \! \times \! 512 \!)$ \!\!\!\!\!} 
 	& min & 0.78 & 0.27 & 0.17 & 0.13 \\
 	& max & 1.51 & 0.33 & 0.34 & 0.17 \\
 	& mean & 0.97 & 0.28 & 0.19 & 0.14 \\ 
 	\hline
	\multirow{3}{*}
	{\!\!\!\!\! peppers \!\!\! $(\!1024 \! \times \! 1024 \!)$ \!\!\!\!\!} 
 	& min & 1.92 & 0.97 & 0.49 & 0.38 \\
 	& max & 2.61 & 1.02 & 0.56 & 0.54 \\
 	& mean & 2.13 & 0.98 & 0.51 & 0.40 \\
 	\hline
	\multirow{3}{*}
	{\!\!\!\!\! peppers \!\!\! $(\!2048 \! \times \! 2048 \!)$ \!\!\!\!\!} 
 	& min & 6.02 & 3.62 & 1.51 & 1.16 \\
 	& max & 6.47 & 3.92 & 1.71 & 1.25 \\
 	& mean & 6.25 & 3.66 & 1.54 & 1.19 \\ 
 	\hline
	\multirow{3}{*}
	{\!\!\!\!\! peppers \!\!\! $(\!4096 \! \times \! 4096 \!)$ \!\!\!\!\!} 
 	& min & 18.20 & 11.49 & 4.15 & 3.33 \\
 	& max & 22.24 & 12.97 & 4.39 & 3.45 \\
 	& mean & 18.79 & 11.59 & 4.19 & 3.37 \\ 
 	\hline     
	\end{tabular}
    \end{small}
\end{table}

\section{Conclusions}
\label{sec:Conclusions}

In this paper, we introduced an optimized parallel UF algorithm for fast CCL using GPUs. 
Our algorithm employs a coarse row-column unification to reduce the computation complexity of local merge and launches a low number of threads for block-to-block connectivity analysis.
As a result, the proposed method can efficiently perform CCL on GPUs in a single scan.
We verified its performance on NVIDIA Geforce GTX 1070 and compared the execution time with those of three other methods.
Experimental results show that the running time of CCL improved greatly compared with the latest method.
The efficiency makes the proposed method suitable for many real-time applications.

%\begin{figure}[t]
%\centering
%\footnotesize
%	\begin{minipage}[b]{0.45\linewidth}
% 		\centering
% 		\subfloat[Local label map after union find]
%		{
% 	 		\begin{overpic}[width=1\textwidth]
% 	 			{fig/localLabelAfterUF.eps}
% 		\end{overpic}
% 	 	}
%	\end{minipage}
%\hskip 2mm
%	\begin{minipage}[b]{0.45\linewidth}
% 		\centering
% 		\subfloat[global label map after union find]
%		{
% 			\begin{overpic}[width=1\textwidth]
% 	 			{fig/globalLabelAfterUF.eps}
% 		\end{overpic}
% 		}
%	\end{minipage}
%\caption{Union find.}
%\label{fig:OpticalModel}
%\vskip -3mm
%\end{figure}
%

\bibliographystyle{IEEEtran}

\bibliography{vcip}

\end{document}